\documentclass{article} 
\usepackage[preprint]{colm2026_conference}

\usepackage{microtype}
\usepackage{hyperref}
\usepackage{url}
\usepackage{booktabs}

\newcommand{\xhdr}[1]{\vspace{1mm}\noindent{{\bf #1.}}}
\usepackage{amsthm}
\usepackage{enumitem}
\usepackage{wrapfig}
\usepackage{booktabs}
\usepackage{multirow}
\usepackage{colortbl}

\usepackage{amsmath}
\usepackage{amssymb}
\usepackage{mathtools}
\usepackage{amsthm}
\usepackage{bbm}
\usepackage{algorithm}
\usepackage{algorithmic}

\usepackage[most]{tcolorbox}
\usepackage{xcolor}
\usepackage{chessboard}
\usepackage{fontawesome5}

\definecolor{CStructural}{HTML}{1F77B4}

\newtcolorbox{ExampleBox}[2][]{%
  enhanced, breakable, sharp corners,
  colframe=#2, colback=#2!2, colbacktitle=#2!12, coltitle=black,
  attach boxed title to top left={xshift=2mm,yshift*=-1.5mm},
  boxed title style={empty, frame code={}, interior code={%
    \fill[#2!12] (frame.south west) rectangle (frame.north east);}},
  title={#1}, fonttitle=\bfseries\sffamily,
  borderline west={3pt}{0pt}{#2}, boxrule=0.6pt,
  left=2mm, right=2mm, top=1.5mm, bottom=1.5mm,
  before skip=6pt, after skip=8pt
}

\newtcolorbox{SubBox}[2][]{%
  enhanced, breakable, sharp corners,
  colframe=#2!60, colback=white, coltitle=black,
  fonttitle=\bfseries\small\sffamily, title={#1},
  boxrule=0.5pt, left=1.2mm, right=1.2mm, top=1.0mm, bottom=1.0mm,
  before skip=1.5mm, after skip=0mm
}

\newcommand{\ExampleBoard}[1]{%
  \centering
  \chessboard[setfen={#1},smallboard]%
  \par\vspace{2pt}%
  {\scriptsize\emph{Black To Move. Board shown for reference only, not included in prompt.}}%
  \par\vspace{4pt}%
}

\usepackage{lineno}

\definecolor{darkblue}{rgb}{0, 0, 0.5}
\hypersetup{colorlinks=true, citecolor=darkblue, linkcolor=darkblue, urlcolor=darkblue}

\title{Grounded Chess Reasoning in Language Models \\ via Master Distillation}

\author{
\bfseries Zhenwei Tang$^{1}$, Qianfeng Wen$^{1}$, Seth Grief-Albert$^{2}$, Yahya Elgabra$^{1}$, \\
\bfseries Blair Yang$^{3}$, Honghua Dong$^{1}$ \& Ashton Anderson$^{1}$ \\[1.2ex]
{\normalfont\small $^{1}$Department of Computer Science, University of Toronto} \\[0.2ex]
{\normalfont\small $^{2}$Queen's University \quad $^{3}$Coolwei AI Lab} \\[0.3ex]
{\normalfont\small Contact: \texttt{\{josephtang, ashton\}@cs.toronto.edu}}
}

\begin{document}

\ifcolmsubmission
\linenumbers
\fi

\maketitle

\begin{center}
\vspace{-10pt}
\href{https://github.com/CSSLab/C1}{\faGithub~\texttt{Code}} \quad
\href{https://huggingface.co/datasets/UofTCSSLab/C1-data}{\raisebox{-0.18em}{\includegraphics[height=1em]{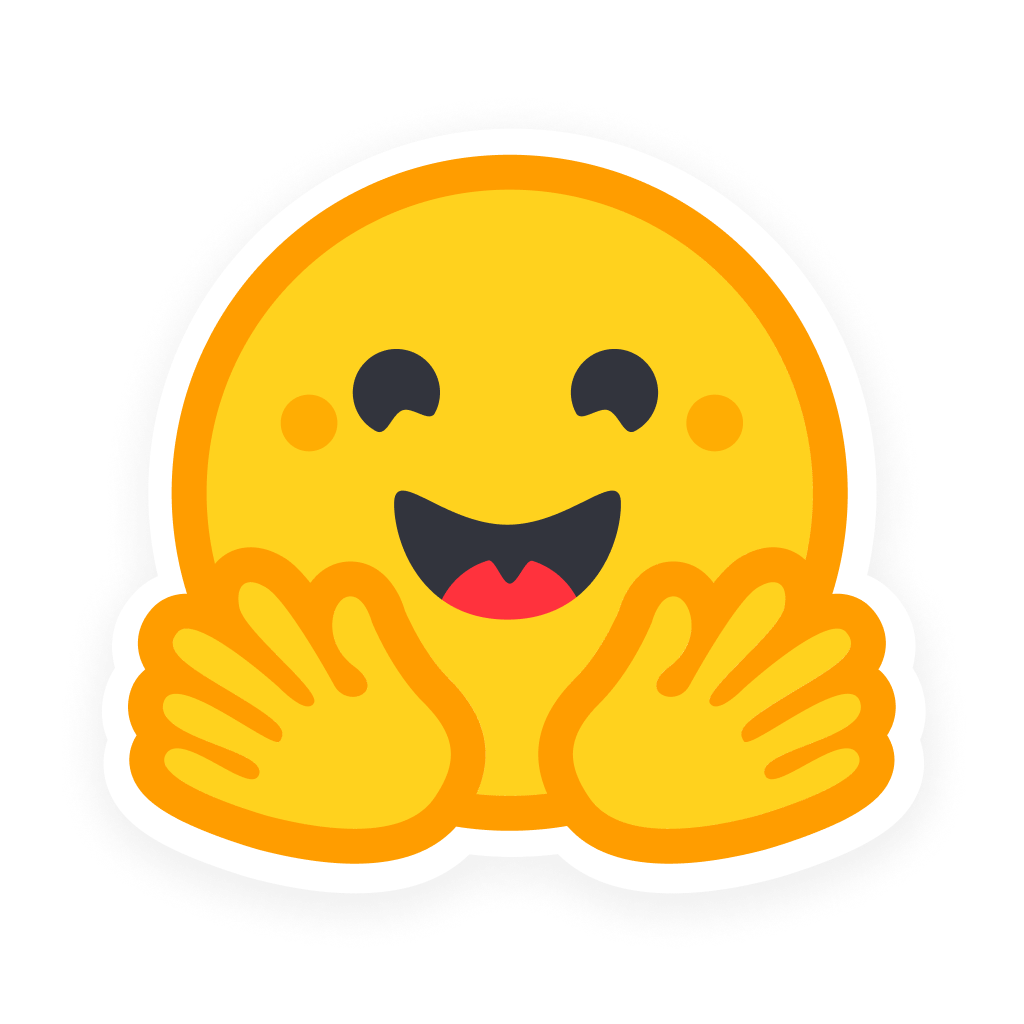}}~\texttt{Dataset}} \quad
\href{https://huggingface.co/UofTCSSLab/C1-4B}{\raisebox{-0.18em}{\includegraphics[height=1em]{figures/hf-logo.png}}~\texttt{Model}}
\vspace{4pt}
\end{center}

\begin{abstract}
Language models often lack grounded reasoning capabilities in specialized domains where training data is scarce but bespoke systems excel. We introduce a general framework for distilling expert system reasoning into natural language chain-of-thought explanations, enabling compact models to acquire domain expertise and the ability to generate faithful, grounded explanations. Rather than distilling only final outputs, we capture the full reasoning process, transforming opaque expert computations into transparent, step-by-step explanations. We demonstrate this approach in chess, a canonical reasoning domain where language models continue to underperform. Our 4B parameter model, C1, advances from a near-zero baseline to 48.1\% accuracy, outperforming all open-source models and most frontier proprietary systems. Notably, C1 surpasses its distillation teacher and generates solutions in two orders of magnitude fewer tokens than baselines. Unlike prior neural chess approaches that predict only best moves, C1 generates explainable solutions revealing strategic reasoning. Our pipeline combines supervised fine-tuning and reinforcement learning with theme-balanced data sampling for comprehensive tactical coverage. Master Distillation demonstrates how to inject expert-level knowledge into compact models for under-optimized domains, offering a recipe for unlocking RLVR where LLMs lack sufficient base capabilities.
\end{abstract}

\section{Introduction}










Large language models exhibit jagged intelligence: superhuman performance in some domains while remaining surprisingly clumsy in others. While the reasons are multifaceted, weak performance often emerges in niche domains where high-quality reasoning data is sparse in pretraining corpora, even when strong specialized systems exist. Chess exemplifies this phenomenon: despite the existence of superhuman engines like Stockfish~\citep{stockfish}, state-of-the-art LLMs struggle with basic tactical puzzles, and recent attempts to apply reinforcement learning with verifiable rewards (RLVR) have failed to yield meaningful improvements~\citep{hwang2025can}.

Prior work has explored distilling chess engines into neural networks, but these approaches transfer only move predictions without the underlying reasoning process~\citep{ruoss2024amortized}. We present Master Distillation, a data synthesis approach that distills both outcome and process from specialized systems into language models. The core insight is that expert systems know the correct answer but cannot articulate their reasoning, while LLMs can generate fluent explanations but lack domain knowledge. Master Distillation combines these complementary strengths: we leverage a master system to provide ground-truth solutions and prompt a frontier LLM to verbalize the reasoning process through Feigned Discovery Prompting, generating chain-of-thought traces that are both verifiably correct and pedagogically natural.

We focus on chess puzzle solving as our testbed. Chess offers perfectly verifiable rewards through engine evaluation, and puzzles have an unambiguous ground truth with a single best move, making them ideal for both evaluation and reward design. The domain is rich with tactical concepts spanning multiple difficulty levels, enabling systematic analysis of model capabilities. Most importantly, the failure of prior RLVR attempts on chess~\citep{hwang2025can} makes it a compelling challenge: if Master Distillation can unlock RLVR here, the approach likely generalizes to other domains where LLMs currently fall short while master systems exist.

Our experiments demonstrate that Master Distillation enables effective SFT+RLVR training for chess reasoning. Our C1-4B model achieves 48.1\% accuracy, outperforming all open-source models and most proprietary models. Notably, C1-4B surpasses its distillation teacher Gemini-3-Flash (40.8\%), demonstrating that the student can exceed teacher performance. Beyond accuracy, our models exhibit remarkable token efficiency, generating solutions in roughly two orders of magnitude fewer tokens than both proprietary and open-source baselines, reflecting how humans actually solve chess puzzles through focused pattern recognition rather than extensive deliberation. In summary, we show that Master Distillation unlocks RLVR for domains where LLMs lack sufficient base capabilities, offering a recipe for improving language model performance in specialized domains with strong bespoke systems. 

\begin{figure*}[t]
\begin{center}
\includegraphics[width=0.9\textwidth]{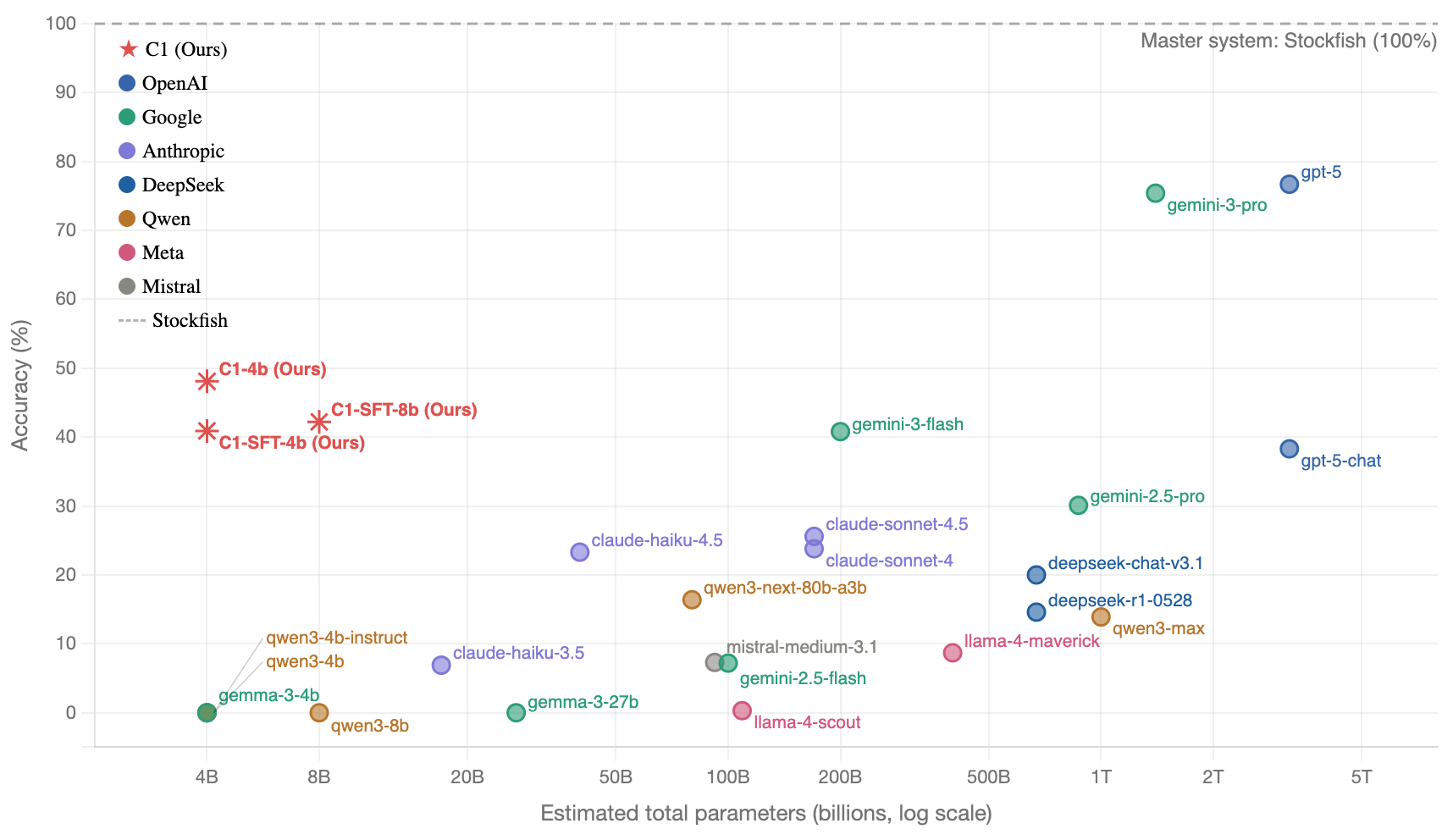}
\end{center}
\caption{C1 achieves competitive performance against frontier models orders of magnitude larger on chess puzzle-solving.} \label{fig:main}
\end{figure*}
\section{Related Work}

\xhdr{Reinforcement Learning with Verifiable Rewards} RLVR enhances language model reasoning by replacing learned reward models with deterministic verification \cite{Guo_2025, hwang2025largelanguagemodelsdevelop, shao2024deepseekmathpushinglimitsmathematical, yu2025dapoopensourcellmreinforcement}, providing objective signals that scale efficiently unlike preference-based RLHF \cite{ouyang2022traininglanguagemodelsfollow}. 
The success of recent reasoning models, including OpenAI o1 \cite{openai2024openaio1card}, DeepSeek-R1 \cite{Guo_2025}, and Qwen3 \cite{yang2025qwen3technicalreport}, demonstrates the effectiveness of RLVR. These methods build on policy gradient foundations like PPO \cite{schulman2017proximalpolicyoptimizationalgorithms}, with GRPO \cite{shao2024deepseekmathpushinglimitsmathematical} marking a critical pivot by eliminating critic networks via grouped rollout baselines. Subsequent work addresses GRPO's limitations: DAPO \cite{yu2025dapoopensourcellmreinforcement} introduces clip-higher bounds and dynamic sampling, GSPO \cite{zheng2025groupsequencepolicyoptimization} stabilizes long sequences via sequence-level ratios, Dr. GRPO \cite{liu2025understandingr1zeroliketrainingcritical} corrects length biases, PRIME \cite{cui2025processreinforcementimplicitrewards} enables implicit process rewards, and RLOO \cite{ahmadian2024basicsrevisitingreinforcestyle} revisits REINFORCE-style optimization.
RLVR has not yet been shown to be effective in chess~\cite{hwang2025largelanguagemodelsdevelop}.

\xhdr{Chess Move Modeling}
Traditional chess AI focuses on finding optimal moves through search and evaluation. AlphaZero~\citep{silver2016mastering, silver2017mastering, silver2017masteringgo} and Leela Chess Zero~\citep{lc0, pascutto2018leela} combine Monte Carlo tree search with neural network evaluation, achieving superhuman performance. Stockfish~\citep{stockfish} integrates Efficiently Updateable Neural Networks (NNUE)~\citep{nasu2018nnue} for fast position evaluation. Recent work explores replacing search with pure neural prediction: \citet{ruoss2024amortized} distill search-based policies into transformers, and \citet{monroe2024mastering} train transformers directly on grandmaster games. These approaches focus on move prediction but do not produce human-readable explanations of their reasoning.
A parallel line of research models human chess behavior rather than optimal play. Maia~\citep{maia} predicts population-level human moves, extended by Maia-2~\citep{maia2} and ALLIE~\citep{allie}. Individual-level modeling has been explored in Maia-Individual~\citep{maia-individual} and Maia4All~\citep{maia4all}. While these models capture human decision-making patterns, they similarly lack explicit reasoning traces.

\xhdr{Chess LLMs}
Several benchmarks evaluate LLM chess capabilities across different dimensions: MATE~\citep{wang2025mate}, PGN2FEN~\citep{cooper2025pgn2fen}, LLM Chess Puzzles~\citep{kagi2025llmchesspuzzles}, LLM Chess Leaderboard~\citep{saplin2025llmchess}, Chess LLM Arena~\citep{jannala2025chessllmarena}, Kaggle Game Arena~\citep{kagglechess}, and ChessQA~\citep{wen2025chessqa}. \citet{kim2025bridging} explores concept-guided chess commentary generation. These benchmarks reveal that current LLMs struggle with chess reasoning despite strong performance in other domains.
Early efforts in training LLMs for chess include commentary generation~\citep{jhamtani2018chesscommentary, zang2019automated}, ChessGPT~\citep{feng2023chessgpt}, and self-supervised approaches using GPT-2~\citep{noever2020chess}. \citet{stockl-2021-watching} analyze language model learning dynamics on chess, while ChePT~\citep{swingle2021chept} combines move prediction with commentary generation. \citet{zhang2025complete} train on complete game transcripts to improve move prediction. However, none of these approaches explicitly address distilling reasoning from master systems like chess engines into language models for explainable puzzle solving, which is the focus of our work.
Most recently, Chess-R1~\citep{hwang2025can} attempted to apply RLVR directly to chess but found that neither direct RLVR nor SFT followed by RLVR yields meaningful improvements. Our work addresses this challenge through Master Distillation, demonstrating that the SFT+RLVR pipeline can be made effective for chess reasoning with a properly designed distillation method.

\section{Methodology}

\subsection{Chess Puzzle Solving with RLVR}
\label{sec:formulation}

\xhdr{Chess Puzzles} Chess puzzles are carefully curated tactical positions with a single objectively best move sequence, designed to test concrete calculation and pattern recognition across various tactical themes such as forks, pins, and mating attacks. Compared to arbitrary game positions where multiple reasonable moves may exist, puzzles offer unambiguous ground truths, making them particularly suitable for evaluation and reward design in our setting.

\xhdr{Chess Puzzle Solving as a Distillation Task} We formulate chess puzzle solving as predicting the correct first move. While this may appear simple, finding the optimal first move requires the model to reason about future positions, anticipate opponent responses, and evaluate resulting outcomes before committing to a decision. This lookahead reasoning is precisely what distinguishes tactical calculation from superficial pattern matching. Prior work on distilling chess engines into neural networks~\citep{ruoss2024amortized,nasu2018nnue} focuses on transferring the engine's move predictions or position evaluations, but not the underlying reasoning process. In contrast, we focus on distilling both the outcome and the process: the model generates a chain-of-thought that explores candidate moves and evaluates their consequences before outputting a final answer. The model is evaluated on first-move accuracy, which serves as a proxy for the quality of its internal reasoning learned from distillation.

\begin{minipage}[c]{1\linewidth}
  \centering
\scalebox{0.8}{
\begin{ExampleBox}[{\faChessKing\ Example Puzzle}]{CStructural}
\begin{minipage}[c]{0.38\linewidth}
  \ExampleBoard{2kr3r/ppp2Npp/2nbp3/6N1/2PP2n1/4B2q/PP2BP2/R2Q1RK1 b - - 2 15}
\end{minipage}%
\hfill
\begin{minipage}[c]{0.58\linewidth}
  {\ttfamily\footnotesize
  You are given a chess position in FEN: 2kr3r/ppp2Npp/2nbp3/6N1/2PP2n1/4B2q/PP2BP2/R2Q1RK1 b - - 2 15. Find the best move for the side to play. PIECE\_ARRANGEMENT. LEGAL\_MOVES. Analyze step by step and explain your reasoning. Finish with a single line formatted EXACTLY as: FINAL\_ANSWER: <answer>. Use UCI notation (e.g., e2e4, c2b1q) for the final answer.
  }
  \textbf{Ground Truth:} {\ttfamily\footnotesize h3h2}
\end{minipage}
\end{ExampleBox}}
\end{minipage}


\xhdr{Established Post Training Pipeline} Reinforcement learning with verifiable rewards (RLVR), with optional supervised fine-tuning (SFT) as a cold-start phase, has emerged as the dominant paradigm for enhancing LLM reasoning capabilities. This pipeline has proven effective across diverse domains: in mathematics, coding, deep research, and tool use~\cite{Guo_2025,yang2025qwen3,jin2025searchr1,li2025websailor}. The success of RLVR relies on a critical prerequisite: the base model must already possess sufficient capability to occasionally produce correct solutions, which RL can then amplify through reward-driven optimization. Chess, however, presents a unique challenge. Despite being a domain with perfectly verifiable rewards, neither direct RLVR nor SFT followed by RLVR yields meaningful improvements~\citep{hwang2025can}. This is because LLMs' base chess capabilities remain too weak to bootstrap from, stemming from the relative scarcity of high-quality chess reasoning data in pretraining corpora. Unlike mathematics, where step-by-step solutions are abundant online, chess analysis rarely includes the detailed chain-of-thought reasoning needed for effective SFT.




\subsection{Master Distillation}

We present a data synthesis approach that combines the domain expertise of specialized systems, e.g., the master, with the linguistic capabilities of frontier LLMs. The core insight is that bespoke systems like Stockfish~\citep{stockfish} know the correct answer but cannot articulate their symbolic reasoning, while LLMs can generate fluent explanations but lack the domain knowledge to reliably solve hard problems.
This motivates our Master Distillation approach: by leveraging a specialized master system to provide ground-truth solutions and a frontier LLM to verbalize the symbolic reasoning process, we can generate high-quality training data that bridges this gap and enables effective RLVR for chess reasoning.




\subsubsection{Context Engineering}
The first dimension that requires careful design towards realizing is context engineering: we must provide the LLM with sufficient information to understand the position and generate grounded analysis.

\xhdr{Master Solution}
The principal variation (PV) is the sequence of optimal moves for both sides leading to the puzzle's objective, which may be checkmate, material gain, or achieving a decisive advantage. The PV represents the symbolic reasoning process of the master system, and is the core knowledge we aim to distill from the symbolic master into the language model student. We leverage Stockfish~\citep{stockfish} at depth 24, which is more than capable of solving most puzzles accurately, as the master system to compute the PV for each puzzle position. We choose not to use transformer-based chess models~\citep{lc0,ruoss2024amortized} because they do not explicitly provide PVs or allow trivial extraction of move sequences. We consider two settings: \textit{multi-PV}, which generates the top-$k$ candidate move sequences and provides additional context about alternative move options, and \textit{best-move PV}, which provides only the single optimal line. While multi-PV offers richer information, best-move PV introduces less noise, allowing the model to focus on what is correct. This is particularly suitable for puzzles where there is one absolute best move and alternatives are significantly worse. We provide empirical comparison between these settings in Table~\ref{tab:data_sft}.

\xhdr{Board and Move Representation}
Chess positions are encoded in Forsyth-Edwards Notation (FEN), a compact string representation that captures piece placement, active color, castling rights, en passant targets, and move clocks~\citep{edwards1994portable}. Moves are represented in Universal Chess Interface (UCI) notation, which specifies moves by their source and destination squares (e.g., e2e4). While Standard Algebraic Notation (SAN) is more human-readable and prevalent in online chess discussions and literature, it introduces ambiguities since the same symbol can refer to different pieces (e.g., B for bishop versus the b-file) and includes special characters like + and \# that complicate evaluation and reward computation through inconsistent string matching. Prior work~\citep{tangseam} has shown that LLMs struggle to recognize FEN strings due to tokenization issues, conflating chess reasoning ability with position recognition. To separate these concerns and provide complete positional context, we augment the FEN with an explicit piece list describing the location of each piece and a list of all legal moves in the current position, for both the puzzle-solving task and distillation. An example chess puzzle is shown in the blue box titled ``Example Puzzle''. 

\xhdr{Additional Hints}
Puzzles may be associated with additional metadata, as provided in public databases like Lichess\footnote{\url{https://database.lichess.org/\#puzzles}}.
We provide the teacher LLM with additional context to guide its analysis. The \textit{opponent's last move} provides crucial context for understanding the tactical situation, as many puzzles arise from the opponent's mistake or create threats that must be addressed. The \textit{tactical themes} (e.g., fork, pin, back-rank mate) hint at the underlying pattern the LLM should explain, helping it focus on the relevant tactical motifs rather than generic analysis. The \textit{puzzle rating} signals the expected complexity, allowing the LLM to calibrate the depth and length of its explanation accordingly. Crucially, while all this information guides the teacher's generation, the student model never sees these hints during training or evaluation.

\subsubsection{Feigned Discovery Prompting}

Our goal is to produce chain-of-thought traces that are both verifiably correct, grounded in the master's solution, and pedagogically natural, reading like genuine reasoning rather than post-hoc rationalization. Naively conditioning on the answer risks generating justifications for a known conclusion rather than authentic problem-solving processes. We want the student model to learn how to arrive at solutions, not merely how to justify them. This creates a fundamental tension between faithfulness to the master trace and authentic reasoning. If the generated reasoning is too faithful, the LLM simply parrots the moves without producing transferable reasoning patterns. If it is too exploratory, the LLM may hallucinate and lose the grounding benefit of the master solution. Our key insight is to instruct the teacher LLM to reason \emph{as if} the solution is unknown, while secretly steering toward the master trace. The teacher must pretend not to know the answer, simulating the discovery process a human solver would experience. This feigned ignorance is critical: it forces the LLM to generate reasoning that explains \emph{why} a move is correct through genuine analysis, rather than simply asserting correctness because the answer was provided. The master's PV and additional context acts as a soft constraint, ensuring the reasoning naturally converges to the correct solution while maintaining the appearance of authentic exploration.

We impose several additional constraints in the prompt to ensure the generated reasoning traces are concise and pedagogically useful. First, we enforce length constraints of 4--10 sentences scaled to puzzle complexity, as chess learners benefit from focused analysis rather than verbose explanations. Second, we require the teacher to bridge chess concepts with notation by annotating moves with natural language clarifications (e.g., ``Qxh7+ (queen takes h7 with check)'') and grounding analysis in explicit board coordinates (e.g., ``the queen on h5'', ``king on g1'') while explaining tactical relationships such as attacks, pins, and defender counts. This bridging is important because student LLMs have limited exposure to chess-specific notation during pretraining. Third, we enforce stylistic constraints: an objective voice without phrases like ``I see'' or ``I notice'', and strict prohibition against mentioning engine scores, ratings, themes, or any indication that the solution was provided. In essence, we are role-playing the LLM as a chess grandmaster who analyzes positions naturally and arrives at the solution through genuine reasoning. These constraints collectively ensure the generated traces read as natural expert analysis rather than machine-generated explanations.

An example of the distillation prompt is shown in the blue box titled ``Master Distillation Prompt'' in the Appendix.

\subsection{C1 Training}

\xhdr{Data Balancing}
We apply theme-balanced sampling to construct both SFT and RLVR training data, as described in Algorithm~\ref{alg:balanced} in the appendix. Since each puzzle is annotated with multiple tactical themes, we prioritize balancing by the rarest themes, ensuring that infrequent concepts such as underPromotion and balestraMate receive sufficient representation. By sampling puzzles that contain these rare themes, more frequent themes like fork and pin are naturally covered as co-occurring labels. We target $M$=800 samples for each of the $K$=50 rarest themes, resulting in approximately 40,000 samples for both SFT and RLVR with non-overlapping puzzle sets. Detailed statistics are provided in Table~\ref{tab:data_stats} and per-theme distributions in Table~\ref{tab:data_stats_theme}.

\xhdr{Supervised Fine-Tuning}
We perform full fine-tuning on Qwen3-4B-Instruct-2507, which achieves near-zero accuracy on our evaluation out of the box. SFT is a necessary prerequisite for RLVR: the base model must solve puzzles correctly some of the time, otherwise all rollouts receive zero reward and no learning signal is available. We opt for full fine-tuning over parameter-efficient methods like LoRA~\cite{hu2022lora} because our task requires knowledge acquisition in a novel domain rather than adaptation of existing capabilities, which benefits from updating all model parameters.

\xhdr{Reinforcement Learning with Verifiable Rewards}
We build upon DAPO~\cite{yu2025dapoopensourcellmreinforcement}, which introduces improvements over GRPO~\cite{shao2024deepseekmathpushinglimitsmathematical}: Overlong Reward Shaping, Token-Level Policy Gradient Loss, Dynamic Sampling, Clip-Higher, and Removing KL Divergence. However, we exclude two of these modifications: Overlong Reward Shaping and Removing KL Divergence. The original DAPO removes KL divergence under the assumption that long-CoT reasoning benefits from allowing the model distribution to diverge significantly from the initial policy. In our setting, the SFT data already consists of concise, to-the-point reasoning traces, and we aim to preserve this brevity rather than encourage free exploration. Similarly, Overlong Reward Shaping is unnecessary since our responses are inherently short. We denote this modified algorithm as DAPO-C1.
We use a binary reward based solely on whether the final move matches the Stockfish-verified optimal move. Following the Bitter Lesson~\cite{sutton2019bitter}, we avoid complex reward shaping that may introduce biases or enable reward hacking, instead relying on a clean signal that scales with data and compute.

\section{Experiments}
\begin{table}[t]
\centering
\vspace{-1.5cm}
\caption{Performance comparison across difficulty levels and models. Avg Acc represents average accuracy (\%), and Avg Tokens indicates average generation length.}
\label{tab:result_split}
\resizebox{\textwidth}{!}{
\begin{tabular}{l||cccc|c||c|cc}
\toprule
 & Beginner & Intermediate & Advanced & Expert & Theme-Split & Avg Acc & Avg Tokens \\
\midrule
\multicolumn{8}{l}{\textit{Proprietary models}} \\
\midrule
\rowcolor{gray!8} gpt-5 & 95.0 & 84.0 & 54.0 & 31.0 & 85.2 & 76.7 & 12,193 \\
\rowcolor{gray!8} gemini-3-pro & 88.0 & 86.0 & 70.0 & 44.0 & 78.2 & 75.4 & 3,182 \\
gemini-3-flash & 65.0 & 59.0 & 34.0 & 19.0 & 38.0 & 40.8 & 6,418 \\
gpt-5-chat & 52.0 & 39.0 & 27.0 & 18.0 & 41.8 & 38.3 & 925 \\
gemini-2.5-pro & 37.0 & 31.0 & 29.0 & 19.0 & 31.0 & 30.1 & 9,668 \\
claude-sonnet-4.5 & 32.0 & 29.0 & 15.0 & 11.0 & 28.6 & 25.6 & 3,227 \\
claude-sonnet-4 & 35.0 & 19.0 & 16.0 & 10.0 & 26.8 & 23.8 & 8,028 \\
claude-haiku-4.5 & 33.0 & 24.0 & 14.0 & 11.0 & 25.6 & 23.3 & 8,111 \\
gemini-2.5-flash & 9.0 & 4.0 & 6.0 & 5.0 & 8.2 & 7.2 & 9,991 \\
\addlinespace[0.1cm]
\midrule
\multicolumn{8}{l}{\textit{Open-source models}} \\
\midrule
deepseek-chat-v3.1 & 27.0 & 21.0 & 6.0 & 16.0 & 22.0 & 20.0 & 11,249 \\
qwen3-next-80b-a3b & 24.0 & 14.0 & 14.0 & 8.0 & 17.6 & 16.4 & 13,938 \\
deepseek-r1-0528 & 11.0 & 10.0 & 14.0 & 16.0 & 16.0 & 14.6 & 14,442 \\
qwen3-max&22.0&15.0&3.0&16.0&13.8&13.9&3,393 \\
llama-4-maverick & 12.0 & 8.0 & 5.0 & 10.0 & 8.6 & 8.7 & 1,092 \\
mistral-medium-3.1 & 9.0 & 6.0 & 7.0 & 4.0 & 8.0 & 7.3 & 2,818 \\
llama-4-scout&0.0&0.0&0.0&1.0&0.4&0.3&806 \\
gemma-3-27b&0.0&0.0&0.0&0.0&0.0&0.0&705 \\ 
\midrule
\addlinespace[0.1cm]
\multicolumn{8}{l}{\textit{Ours}} \\
\midrule
C1-SFT-4B & 51.0 & 30.0 & 30.0 & 26.0 & 46.2 & 40.9 & 188 \\
C1-SFT-8B & 57.0 & 36.0 & 27.0 & 27.0 & 46.6 & 42.2 & 189 \\
\rowcolor{gray!20} C1-4B & 65.0 & 39.0 & 39.0 & 22.0 & 53.6 & 48.1 & 178 \\
\bottomrule
\end{tabular}}
\end{table}

\subsection{Experimental Settings}

We implement SFT using LLaMA-Factory~\citep{zheng2024llamafactory} and RLVR using VERL~\citep{sheng2024hybridflow}. All experiments are conducted on 4$\times$H100 GPUs. For both SFT and RLVR, we use a held-out validation set of chess puzzles for early stopping to prevent overfitting. Detailed hyperparameters for SFT and RLVR are provided in Tables \ref{tab:rlvr_hyperparams} and \ref{tab:sft_hyperparams} in the Appendix, respectively. We follow the evaluation protocol of~\citet{wen2025chessqa}. The test set contains 900 puzzles: 500 puzzles sampled across 20 tactical themes (25 per theme) and 400 puzzles sampled across 4 difficulty levels (100 per level: Beginner, Intermediate, Advanced, Expert). We report pass@1 accuracy as the primary metric. For all baseline models, we enable reasoning mode where applicable. We use \href{https://openrouter.ai/}{OpenRouter} for API access to proprietary models during evaluation. We use Gemini-3-Flash-Preview via OpenRouter as the teacher model for Master Distillation.

\subsection{Performance Comparison}



Our C1-4B model achieves the highest accuracy among all open-source models by a substantial margin and outperforms most frontier proprietary models, demonstrating the effectiveness of Master Distillation for domain-specific reasoning.
Notably, all C1 models surpass Gemini-3-Flash, the model used to generate these reasoning traces, indicating that combining symbolic expertise with language model reasoning enables the student to exceed its teacher. This improvement is consistent across difficulty levels, with particularly strong gains on Intermediate puzzles. On the Theme-Split evaluation, C1-4B shows even more pronounced advantages, with detailed per-theme results provided in Appendix Table~\ref{tab:result_theme}.

Beyond accuracy, our models exhibit remarkable token efficiency, generating solutions in roughly two orders of magnitude fewer tokens than both proprietary and open-source baselines. This reflects how humans actually solve chess puzzles: through focused pattern recognition and precise calculation rather than extensive deliberation. It also aligns with the practical needs of chess learners, who benefit from concise, to-the-point explanations rather than verbose reasoning chains.

Comparing C1-4B (SFT+RLVR) with C1-SFT-4B reveals the effect of reinforcement learning. RLVR provides consistent improvements on Beginner and Intermediate puzzles, but shows diminished or negative gains on Expert-level problems. This pattern suggests that the chain-of-thought reasoning traces may not faithfully capture the underlying logic for harder puzzles, limiting what RLVR can learn from outcome-based rewards. Unlike mathematical reasoning, where RLVR often yields dramatic improvements, the less reliable reasoning chains in complex chess puzzles constrain the potential gains from reinforcement learning.

\begin{table}[t]
\centering
\vspace{-0.5cm}
\caption{Ablation study on SFT data configurations.}
\label{tab:data_sft}
\setlength{\tabcolsep}{2.5pt}
\resizebox{\textwidth}{!}{
\begin{tabular}{@{}l||cc|c|cc|ccc||>{\columncolor{gray!20}}c@{}}
    \toprule
    Scale   & 8k & 8k & 8k & 8k & 16k & 8k & 8k & 8k & 39k \\
    Distribution & random & hard & balanced & balanced & balanced & balanced & balanced & balanced & balanced \\
    Quality & flash & flash & pro & flash & flash & flash & flash & flash & flash \\
    Context & full & full & full & full & full & Multi PVs & w/o Theme & w/o Feigned & full \\
    \midrule
    SFT Results & 19.3 & 16.2 & 22.8 & 20.1 & 29.7 & 17.6 & 17.3 & 16.3 & 40.9 \\
    \bottomrule
\end{tabular}}
\end{table}

\subsection{Additional Findings}

\xhdr{SFT Data} Table~\ref{tab:data_sft} presents ablation studies on SFT data configurations, revealing several key insights. First, distillation quality matters: using Gemini-3-Pro as the teacher outperforms Gemini-3-Flash, suggesting that higher-quality chain-of-thought traces lead to better student performance. However, due to budget constraints, we use Flash for scaling experiments. Second, data distribution significantly impacts results: hard sampling yields the worst performance because chain-of-thought quality degrades on difficult puzzles, while random sampling underperforms balanced sampling due to insufficient coverage of rare tactical concepts. Third, our context engineering and prompting design choices are validated: Multi-PV introduces noise and performs worse than best-move PV, removing theme hints causes the reasoning to lose focus, and removing our Feigned Discovery Prompt (w/o Feigned) leads to the largest degradation, demonstrating that our prompting strategy is critical for generating effective training data. Fourth, data scale matters substantially, with performance improving from 19.3\% at 8k to 29.7\% at 16k and 40.9\% at 39k samples. These results collectively justify our final configuration (highlighted in gray): 39k balanced samples distilled from Gemini-3-Flash with full context and Feigned Discovery Prompting.

\begin{table}[t!]
  \centering
  \vspace{-0.5cm}
  \caption{Ablation study on RLVR data and SFT configurations.}
  \begin{minipage}{0.55\textwidth}
    \centering
    \begin{tabular}{l l | c c c}
        \toprule
        \textbf{Distribution} & \textbf{Reuse} & \textbf{SFT} & \textbf{RLVR} & \textbf{Impr} \\
        \midrule
        random & false & 40.9 & 47.1 & +6.2 \\
        hard & false & 40.9 & 43.0 & +1.1 \\
        balanced & true & 40.9 & 46.7 & +5.8 \\
        \rowcolor{gray!20}balanced & false & 40.9 & 48.1 & +7.2 \\
        \bottomrule
    \end{tabular}
    \label{tab:rlvr_data}
  \end{minipage}%
  \hfill
  \begin{minipage}{0.45\textwidth}
    \centering
    \begin{tabular}{l c}
        \toprule
        \textbf{Configuration} & \textbf{SFT} \\
        \midrule
        LoRA ($r=16, \alpha=32$) & 32.1 \\
        LoRA ($r=64, \alpha=128$) & 38.8 \\
        \rowcolor{gray!20}Full Fine-Tuning & 40.9 \\
        \bottomrule
    \end{tabular}
    \label{tab:peft}
  \end{minipage}
  \vspace{-0.5cm}
\end{table}


\xhdr{RLVR Data} Table~\ref{tab:rlvr_data} (left) presents ablations on RLVR data selection strategies. Random sampling draws problems i.i.d. from the training distribution, while hard sampling prioritizes the most difficult puzzles. Balanced sampling ensures equal representation across tactical themes for broader concept coverage. The results show that balanced sampling achieves the best performance, suggesting that diverse tactical exposure benefits reinforcement learning more than focusing on difficulty alone. Notably, hard sampling yields the smallest improvement, consistent with our earlier observation that unfaithful reasoning traces on difficult puzzles limit RLVR's effectiveness.
We also examine whether to \textit{reuse} SFT samples during RLVR. As shown in~\ref{tab:rlvr_data}, reusing SFT samples slightly hurts performance compared to using fresh problems. This is likely because training on previously seen examples encourages the model to reinforce existing behavior rather than acquire new knowledge, shifting the learning objective toward consistency with prior outputs rather than exploration of improved reasoning strategies.



\xhdr{SFT Training} Table~\ref{tab:peft} (right) validates our choice of full fine-tuning over parameter-efficient methods. While increasing LoRA rank from 16 to 64 improves performance from 32.1\% to 38.8\%, full fine-tuning achieves the best result at 40.9\%. This confirms our hypothesis that chess puzzle solving requires knowledge acquisition in a novel domain rather than adaptation of existing capabilities, benefiting from updating all model parameters rather than low-rank approximations.

\xhdr{RLVR Training} Table~\ref{tab:algo_ablations} (left) presents ablation studies on RLVR configurations. DAPO outperforms GRPO, validating the benefits of Token-Level Policy Gradient Loss, Dynamic Sampling, and Clip-Higher for our setting. Our modified DAPO-C1, which retains KL divergence and removes Overlong Reward Shaping to preserve concise reasoning, achieves the best performance with a 7.2\% improvement over SFT. We also experiment with partial credit rewards, where moves with correct source or destination squares receive a fractional reward $\eta$. However, partial rewards with $\eta$=0.5 hurts performance, and $\eta$=0.2 yields only marginal gains. This supports our design choice of using binary correctness rewards: following the Bitter Lesson~\citep{sutton2019bitter}, simple and clean reward signals scale better than hand-crafted shaping, which may introduce biases or enable reward hacking. Our final configuration (highlighted in gray) uses DAPO-C1 with binary correctness rewards.



\begin{table}[t!]
  \centering
  \vspace{-1cm}
  \caption{Ablation study on RLVR configurations and base models. Qwen3-8B was not trained with RLVR due to computational constraints. Qwen3-4B-it denotes Qwen3-4B-Instruct-2507.}
  \begin{minipage}[t]{0.4\textwidth}
    \centering
    \label{tab:algo_ablations}
    \small
    \begin{tabular}{l l | c c c}
        \toprule
        \textbf{Algorithm} & \textbf{Reward} & \textbf{SFT} & \textbf{RLVR} & \textbf{Impr} \\
        \midrule
        GRPO & Correctness & 40.9 & 45.5 & +4.6 \\
        DAPO & Correctness & 40.9 & 47.3 & +6.4 \\
        \midrule
        DAPO-C1 & Partial ($\eta$=0.5) & 40.9 & 39.3 & -1.6 \\
        DAPO-C1 & Partial ($\eta$=0.2) & 40.9 & 41.6 & +0.7 \\
        \midrule
        \rowcolor{gray!20}DAPO-C1 & Correctness & 40.9 & 48.1 & +7.2 \\
        \bottomrule
    \end{tabular}
  \end{minipage}%
  \hfill
  \begin{minipage}[t]{0.42\textwidth}
    \centering
    \label{tab:models}
    \begin{tabular}{l |c c}
        \toprule
        \textbf{Model} & \textbf{SFT} & \textbf{RLVR} \\
        \midrule
        Qwen3-8B & 42.2 & \text{-} \\
        Qwen3-4B & 40.5 & 47.6 \\
        \rowcolor{gray!20}Qwen3-4B-it & 40.9 & 48.1 \\
        Gemma3-4B & 36.2 & 41.0 \\
        \bottomrule
    \end{tabular}
  \end{minipage}
  \vspace{-0.5cm}
\end{table}




\xhdr{Model} Table~\ref{tab:models} (right) compares different base models after SFT and RLVR training. Qwen3-4B-Instruct-2507 achieves the best overall performance, slightly outperforming the reasoning variant Qwen3-4B, possibly because reasoning models are more rigid in their established reasoning patterns and thus harder to adapt to domain-specific chain-of-thought styles. Gemma3-4B lags behind the Qwen models, reflecting differences in pretraining data and architecture. Notably, the 8B model shows only marginal gains over the 4B model after SFT (42.2\% vs 40.9\%), indicating that data quality is a more significant bottleneck than model capacity at this scale. This suggests that simply scaling parameters without improving training data yields diminishing returns for domain-specific reasoning tasks. While scaling to larger models would likely yield further improvements, we focus on the 4B model due to computational constraints and leave scaling experiments to future work.

\section{Discussion}

\xhdr{Towards Better Performance}
Our ablation studies reveal several key factors for effective chess reasoning: high-quality chain-of-thought traces from capable teacher models, balanced data distribution for broad tactical coverage, careful context engineering and Feigned Discovery Prompting, and sufficient data scale. Each of these dimensions offers opportunities for improvement.
The most straightforward path to better performance is scaling. Our experiments show consistent gains from 8k to 16k to 39k training samples, with no sign of saturation. Similarly, using Gemini-3-Pro instead of Flash as the teacher improves performance, suggesting that even stronger teachers would yield higher-quality distillation data. Notably, our distilled C1 models surpass their teacher Gemini-3-Flash, demonstrating that Master Distillation combined with RLVR can produce students that exceed teacher performance. Due to resource constraints, we focused on 4B parameter models, but our methodology naturally extends to larger architectures where we would expect further gains.

\xhdr{Chess Foundational Model}
While C1 performs well on puzzles with unambiguous ground truth, a true chess foundation model must generalize to strategic positions where multiple reasonable continuations exist and evaluation is less binary. This requires rethinking reward design for Master Distillation. Beyond move prediction, such a model could support game annotation, opening preparation, endgame analysis, mistake identification, and personalized training. Mid-training on large-scale chess corpora with multimodal capabilities to process board images could further enhance utility. The ultimate goal is accessible chess education: unlike engines that provide optimal but opaque suggestions beyond human comprehension, a foundation model trained with Master Distillation could explain concepts at appropriate skill levels and guide learners through tactical and strategic themes, democratizing high-quality chess instruction.

\xhdr{Master Distillation Generalization}
Master Distillation applies to any domain where strong bespoke systems exist but LLMs lack reasoning capabilities due to sparse training data. The key requirements are a master system providing verifiable solutions and problems with clear correctness criteria. Examples include circuit design (simulation-validated), music composition (harmonic/counterpoint verification), and molecular docking (binding affinity scoring), all domains where automated verification exists but LLM-generated reasoning remains limited. While the master system component transfers directly, Feigned Discovery Prompting requires domain-specific redesign: the core principle of the LLM pretending to discover the solution through genuine reasoning remains constant, but what constitutes authentic reasoning varies by domain.

\clearpage
\bibliography{colm2026_conference}
\bibliographystyle{colm2026_conference}

\newpage
\appendix
\begin{figure*}[htbp]
\begin{center}
\includegraphics[width=0.95\textwidth]{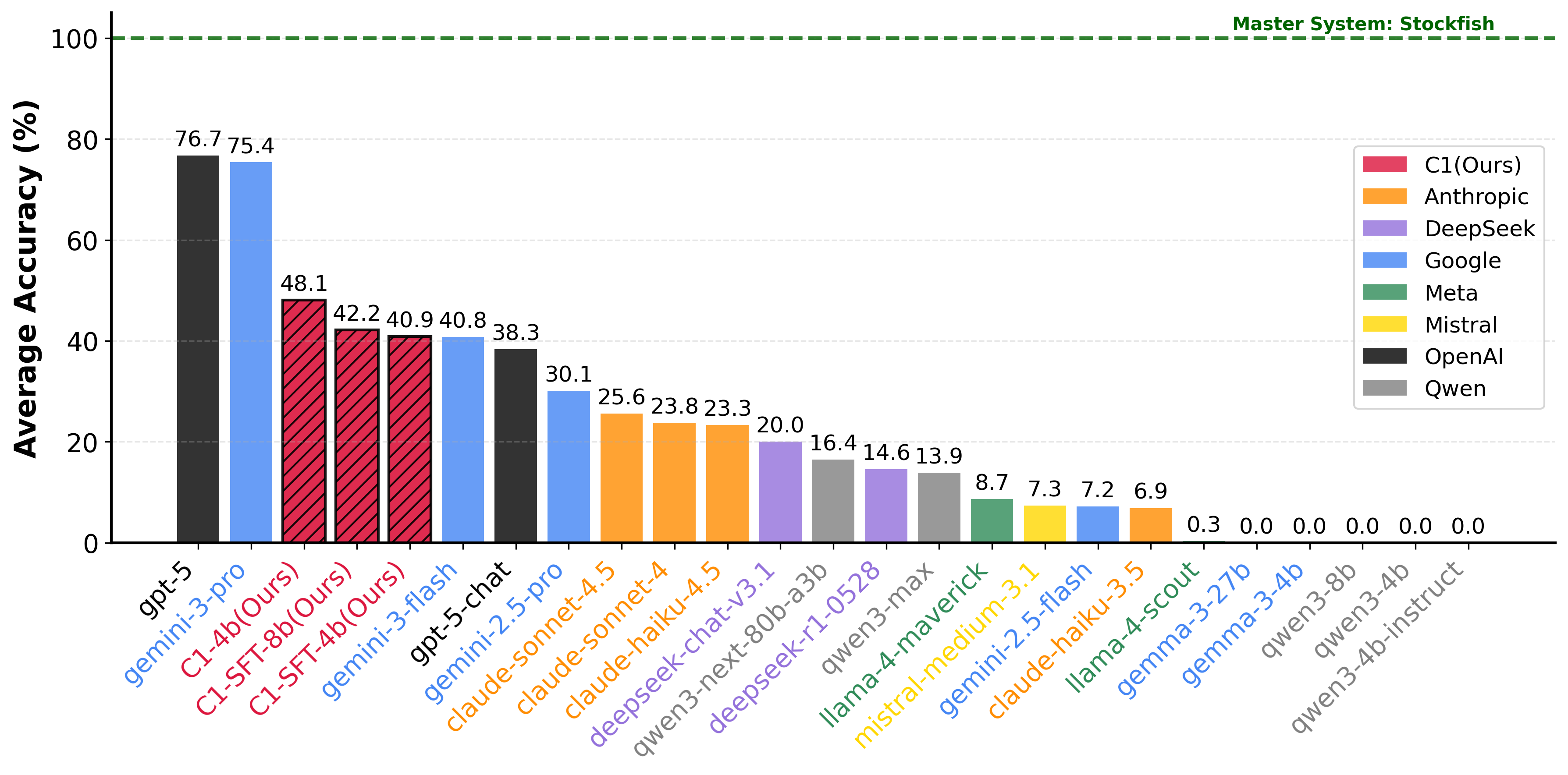}
\end{center}
\caption{Chess puzzle-solving accuracy of C1 compared to frontier LLMs.} \label{fig:main_sub}
\end{figure*}

\begin{ExampleBox}[{\faChessKing\ Master Distillation Prompt}]{CStructural}
\begin{tcbraster}[raster columns=1, raster equal height=rows, raster column skip=2.5mm]
  \begin{SubBox}[General Instruction]{CStructural}
    {\ttfamily\footnotesize
    You are a chess grandmaster generating training data for teaching small LLMs to solve chess puzzles.
    
    CRITICAL: The small model only sees the FEN string. Your analysis must show how to go from FEN → piece positions → tactical relationships → solution. Ground everything in explicit square names.
    }
  \end{SubBox}
  \begin{SubBox}[Context (provided to teacher, hidden from student)]{CStructural}
    {\ttfamily\footnotesize
    FEN: \{fen\}\\
    Pieces: \{piece\_arrangement\}\\
    Side to Move: \{side\_to\_move\}\\
    Opponent's Last Move: \{last\_move\}\\
    Solution: \{move\}\\
    Rating: \{rating\}\\
    Themes: \{themes\}\\
    PVs: \{principal\_variation\} (omitted for mateIn1) \\
    Legal Moves: \{legal\_moves\}
    }
  \end{SubBox}
  \begin{SubBox}[Task Description]{CStructural}
    {\ttfamily\footnotesize
    TASK: Write natural chain-of-thought analysis arriving at \{move\}.
    
    Your analysis should cover (in natural prose, vary the structure):
    \begin{itemize}[leftmargin=*, nosep]
      \item Where key pieces are (use square names: ``the queen on h5'', ``king on g1'')
      \item What tactical relationship exists (attacks, pins, weak squares, defender counts)
      \item Why the move works (what it threatens, why opponent can't respond adequately)
    \end{itemize}
    
    Style:
    \begin{itemize}[leftmargin=*, nosep]
      \item Objective voice, no ``I see/notice''
      \item Standard notation with brief clarification when helpful: ``Qxh7+ (queen takes h7 with check)''
      \item Never mention engine scores, ratings, themes, or that you were given the solution
      \item 4--10 sentences, scaled to complexity
    \end{itemize}
    
    End with exactly: FINAL\_ANSWER: \{move\}
    }
  \end{SubBox}
\end{tcbraster}
\end{ExampleBox}

\begin{table}[h]                                                                                       
\centering                                                                                             
\caption{RLVR Training Hyperparameters.}                                                    
\label{tab:rlvr_hyperparams}                                                                           
\begin{tabular}{ll}                                                                                    
\toprule                                                                                               
\textbf{Hyperparameter} & \textbf{Value} \\                                                            
\midrule                                                                                               
\multicolumn{2}{l}{\textit{Model Configuration}} \\                                                    
Base Model & C1-SFT-4B \\                                                
Algorithm & DAPO-C1 \\                                               
Gradient Checkpointing & Enabled \\                                                                    
\midrule                                                                                               
\multicolumn{2}{l}{\textit{Data Configuration}} \\                                                                                                       
Max Prompt Length & 1024 \\                                                                            
Max Response Length & 512 \\                                                                           
Generation Batch Size & 96 \\                                                                          
Training Batch Size & 32 \\                                                                            
Rollout Samples & 32 \\                                                                          
\midrule                                                                                               
\multicolumn{2}{l}{\textit{Sampling Parameters}} \\                                                    
Temperature & 1.0 \\                                                                                   
Top-$p$ & 1.0 \\                                                                                       
Top-$k$ & -1 \\                                                                         
GPU Memory Utilization & 0.8 \\                                                                        
\midrule                                                                                               
\multicolumn{2}{l}{\textit{Optimizer}} \\                                                              
Learning Rate & $1 \times 10^{-6}$ \\                                                                  
PPO Mini Batch Size & 8 \\                                                                             
PPO Micro Batch Size (per GPU) & 32 \\                                                                 
Log Prob Micro Batch Size & 256 \\                                                                     
\midrule                                                                                               
\multicolumn{2}{l}{\textit{RL Specific}} \\                                                                                                              
KL Loss Coefficient & 0.001 \\                                                                                  
Entropy Coefficient & 0 \\                                                                             
Clip Ratio (low / high) & 0.2 / 0.28 \\                                                                
Loss Aggregation & Token-mean \\                                                                       
KL in Reward & Disabled \\                                                                                                                      
Filter Groups & Enabled (max batches: 5) \\                                             
\midrule                                                                                               
\multicolumn{2}{l}{\textit{Training}} \\                                                               
Total Epochs & 1 \\                                                                                    
Save Frequency & 500 \\                                                                                
Test Frequency & 10 \\                                                                                 
Num GPUs & 4 \\   
Early-Stop Tolerance & 5 \\
\bottomrule                                                                                            
\end{tabular}                                                                                          
\end{table}

\begin{table*}[htbp]                                                                                      
\centering                                                                                             
\caption{Supervised Fine-Tuning Hyperparameters}                                                       
\label{tab:sft_hyperparams}                                                                            
\begin{tabular}{ll}                                                                                    
\toprule                                                                                               
\textbf{Parameter} & \textbf{Setting} \\                                                               
\midrule                                                                                               
Base Model & Qwen3-4B-Instruct \\                                                                      
Training Method & Full-parameter fine-tuning \\                                                        
Optimizer & AdamW (DeepSpeed ZeRO-3) \\                                                                
Learning Rate & $1\mathrm{e}{-5}$ \\                                                                   
Batch Size & $16 \times 4$ (grad accum) $\times$ 4 GPUs = 256 \\                                       
Epochs & 10 \\                                                                                         
Scheduler & Cosine with 10\% warmup \\                                                                 
Max Length & 1024 tokens \\                                                                            
Precision & BF16 \\ 
Validation Interval & 50 Steps \\
Early-Stop Tolerance & 5 \\
\bottomrule                                                                                            
\end{tabular}                                                                                          
\end{table*}

\begin{table}[htbp]
\centering
\caption{Data statistics for SFT and RLVR training.}
\label{tab:data_stats}
\begin{tabular}{lcc}
\toprule
& \textbf{SFT} & \textbf{RLVR} \\
\midrule
Total samples & 39,609 & 39,585 \\
Unique themes & 66 & 66 \\
Balanced themes ($K$) & 50 & 50 \\
Target per theme ($M$) & 800 & 800 \\
Rating range & 399--3,154 & 399--3,196 \\
Average rating & 1,540 & 1,539 \\
\bottomrule
\end{tabular}
\end{table}

\begin{table*}[t]
\small
\centering
\caption{Theme distribution in SFT and RLVR training data. We apply balanced sampling with $K=50$ rare themes and target $M=800$ samples per theme.}
\label{tab:data_stats_theme}
\small
\resizebox{\textwidth}{!}{
\begin{tabular}
{l||c@{\hspace{6pt}}c@{\hspace{6pt}}c@{\hspace{6pt}}c@{\hspace{6pt}}c@{\hspace{6pt}}c@{\hspace{6pt}}c@{\hspace{6pt}}c@{\hspace{6pt}}c@{\hspace{6pt}}c@{\hspace{6pt}}c@{\hspace{6pt}}c}
\toprule
 & \rotatebox{90}{endgame} & \rotatebox{90}{mate} & \rotatebox{90}{short} & \rotatebox{90}{middlegame} & \rotatebox{90}{crushing} & \rotatebox{90}{long} & \rotatebox{90}{advantage} & \rotatebox{90}{mateIn2} & \rotatebox{90}{oneMove} & \rotatebox{90}{mateIn1} & \rotatebox{90}{master} & \rotatebox{90}{veryLong} \\
\midrule
SFT & 20,817 & 17,971 & 17,205 & 16,210 & 12,957 & 10,236 & 7,858 & 6,930 & 6,856 & 6,824 & 6,133 & 5,294 \\
RLVR & 20,752 & 17,872 & 17,056 & 16,210 & 13,054 & 10,292 & 7,834 & 6,767 & 6,860 & 6,833 & 6,056 & 5,359 \\
\bottomrule
\end{tabular}}

\vspace{0.3cm}

\resizebox{\textwidth}{!}{
\begin{tabular}
{l||c@{\hspace{6pt}}c@{\hspace{6pt}}c@{\hspace{6pt}}c@{\hspace{6pt}}c@{\hspace{6pt}}c@{\hspace{6pt}}c@{\hspace{6pt}}c@{\hspace{6pt}}c@{\hspace{6pt}}c@{\hspace{6pt}}c@{\hspace{6pt}}c@{\hspace{6pt}}c@{\hspace{6pt}}c@{\hspace{6pt}}c@{\hspace{6pt}}c}
\toprule
 & \rotatebox{90}{sacrifice} & \rotatebox{90}{advancedPawn} & \rotatebox{90}{defensiveMove} & \rotatebox{90}{kingsideAttack} & \rotatebox{90}{fork} & \rotatebox{90}{attraction} & \rotatebox{90}{opening} & \rotatebox{90}{pin} & \rotatebox{90}{mateIn3} & \rotatebox{90}{quietMove} & \rotatebox{90}{rookEndgame} & \rotatebox{90}{discoveredAttack} & \rotatebox{90}{deflection} & \rotatebox{90}{exposedKing} & \rotatebox{90}{pawnEndgame} & \rotatebox{90}{promotion} \\
\midrule
SFT & 4,932 & 3,288 & 3,277 & 3,001 & 2,948 & 2,672 & 2,572 & 2,370 & 2,335 & 2,307 & 2,195 & 2,164 & 2,136 & 2,019 & 1,829 & 1,820 \\
RLVR & 5,037 & 3,316 & 3,178 & 2,957 & 2,939 & 2,718 & 2,614 & 2,364 & 2,395 & 2,278 & 2,159 & 2,245 & 2,279 & 2,005 & 1,850 & 1,901 \\
\bottomrule
\end{tabular}}

\vspace{0.3cm}

\resizebox{\textwidth}{!}{
\begin{tabular}
{l||c@{\hspace{6pt}}c@{\hspace{6pt}}c@{\hspace{6pt}}c@{\hspace{6pt}}c@{\hspace{6pt}}c@{\hspace{6pt}}c@{\hspace{6pt}}c@{\hspace{6pt}}c@{\hspace{6pt}}c@{\hspace{6pt}}c@{\hspace{6pt}}c@{\hspace{6pt}}c@{\hspace{6pt}}c@{\hspace{6pt}}c@{\hspace{6pt}}c@{\hspace{6pt}}c@{\hspace{6pt}}c}
\toprule
 & \rotatebox{90}{backRankMate} & \rotatebox{90}{masterVsMaster} & \rotatebox{90}{hangingPiece} & \rotatebox{90}{queensideAttack} & \rotatebox{90}{skewer} & \rotatebox{90}{clearance} & \rotatebox{90}{doubleCheck} & \rotatebox{90}{mateIn4} & \rotatebox{90}{intermezzo} & \rotatebox{90}{queenEndgame} & \rotatebox{90}{bishopEndgame} & \rotatebox{90}{queenRookEndgame} & \rotatebox{90}{zugzwang} & \rotatebox{90}{attackingF2F7} & \rotatebox{90}{trappedPiece} & \rotatebox{90}{knightEndgame} & \rotatebox{90}{mateIn5} & \rotatebox{90}{capturingDefender} \\
\midrule
SFT & 1,596 & 1,557 & 1,439 & 1,393 & 1,165 & 1,083 & 1,077 & 1,055 & 1,040 & 966 & 964 & 891 & 874 & 872 & 867 & 866 & 827 & 827 \\
RLVR & 1,594 & 1,587 & 1,515 & 1,425 & 1,122 & 1,081 & 1,061 & 1,047 & 1,053 & 967 & 973 & 898 & 892 & 871 & 874 & 850 & 830 & 819 \\
\bottomrule
\end{tabular}}

\vspace{0.3cm}

\resizebox{\textwidth}{!}{
\begin{tabular}
{l||c@{\hspace{6pt}}c@{\hspace{6pt}}c@{\hspace{6pt}}c@{\hspace{6pt}}c@{\hspace{6pt}}c@{\hspace{6pt}}c@{\hspace{6pt}}c@{\hspace{6pt}}c@{\hspace{6pt}}c@{\hspace{6pt}}c@{\hspace{6pt}}c@{\hspace{6pt}}c@{\hspace{6pt}}c@{\hspace{6pt}}c@{\hspace{6pt}}c@{\hspace{6pt}}c@{\hspace{6pt}}c@{\hspace{6pt}}c@{\hspace{6pt}}c}
\toprule
 & \rotatebox{90}{xRayAttack} & \rotatebox{90}{cornerMate} & \rotatebox{90}{interference} & \rotatebox{90}{arabianMate} & \rotatebox{90}{hookMate} & \rotatebox{90}{anastasiaMate} & \rotatebox{90}{equality} & \rotatebox{90}{vukovicMate} & \rotatebox{90}{blindSwineMate} & \rotatebox{90}{triangleMate} & \rotatebox{90}{superGM} & \rotatebox{90}{enPassant} & \rotatebox{90}{killBoxMate} & \rotatebox{90}{castling} & \rotatebox{90}{bodenMate} & \rotatebox{90}{doubleBishopMate} & \rotatebox{90}{dovetailMate} & \rotatebox{90}{smotheredMate} & \rotatebox{90}{balestraMate} & \rotatebox{90}{underPromotion} \\
\midrule
SFT & 825 & 818 & 817 & 810 & 808 & 806 & 805 & 805 & 805 & 805 & 803 & 803 & 802 & 800 & 800 & 800 & 800 & 800 & 697 & 512 \\
RLVR & 830 & 822 & 814 & 807 & 810 & 808 & 807 & 808 & 805 & 808 & 802 & 805 & 800 & 800 & 800 & 801 & 801 & 801 & 668 & 517 \\
\bottomrule
\end{tabular}}
\end{table*}

\begin{table*}[h]
\centering
\caption{Performance comparison across chess tactic themes. All values represent accuracy.}
\small
\label{tab:result_theme}
\resizebox{\textwidth}{!}{
\begin{tabular}
{l||c@{\hspace{8pt}}c@{\hspace{8pt}}c@{\hspace{8pt}}c@{\hspace{8pt}}c@{\hspace{8pt}}c@{\hspace{8pt}}c@{\hspace{8pt}}c@{\hspace{8pt}}c@{\hspace{8pt}}c@{\hspace{8pt}}c@{\hspace{8pt}}c@{\hspace{8pt}}c@{\hspace{8pt}}c@{\hspace{8pt}}c@{\hspace{8pt}}c@{\hspace{8pt}}c@{\hspace{8pt}}c@{\hspace{8pt}}c@{\hspace{8pt}}c}
\toprule
 & \rotatebox{90}{advancedPawn} & \rotatebox{90}{attraction} & \rotatebox{90}{backRankMate} & \rotatebox{90}{capturingDefender} & \rotatebox{90}{defensiveMove} & \rotatebox{90}{deflection} & \rotatebox{90}{discoveredAttack} & \rotatebox{90}{doubleCheck} & \rotatebox{90}{fork} & \rotatebox{90}{hangingPiece} & \rotatebox{90}{mateIn1} & \rotatebox{90}{mateIn2} & \rotatebox{90}{pin} & \rotatebox{90}{promotion} & \rotatebox{90}{queensideAttack} & \rotatebox{90}{sacrifice} & \rotatebox{90}{skewer} & \rotatebox{90}{trappedPiece} & \rotatebox{90}{xRayAttack} & \rotatebox{90}{zugzwang} \\
\midrule
\multicolumn{21}{l}{\textit{Proprietary models}} \\
\midrule
\rowcolor{gray!8} gpt-5 & 84 & 92 & 100 & 84 & 68 & 92 & 80 & 80 & 88 & 96 & 96 & 96 & 84 & 76 & 92 & 92 & 84 & 68 & 80 & 72 \\
\rowcolor{gray!8} gemini-3-pro & 80 & 68 & 92 & 80 & 68 & 68 & 84 & 72 & 88 & 80 & 84 & 84 & 76 & 68 & 96 & 88 & 84 & 56 & 80 & 68 \\
gemini-3-flash & 28 & 32 & 32 & 44 & 40 & 44 & 36 & 20 & 40 & 44 & 56 & 52 & 60 & 24 & 44 & 52 & 36 & 24 & 28 & 24 \\
gpt-5-chat & 28 & 40 & 68 & 36 & 44 & 48 & 32 & 36 & 48 & 32 & 44 & 64 & 32 & 40 & 44 & 52 & 48 & 28 & 40 & 32 \\
gemini-2.5-pro & 28 & 20 & 52 & 8 & 36 & 20 & 20 & 52 & 32 & 36 & 40 & 28 & 28 & 44 & 36 & 28 & 32 & 28 & 20 & 32 \\
claude-sonnet-4.5 & 44 & 12 & 52 & 16 & 8 & 20 & 48 & 40 & 48 & 8 & 16 & 36 & 20 & 56 & 12 & 40 & 32 & 0 & 40 & 24 \\
claude-sonnet-4 & 36 & 28 & 44 & 16 & 16 & 28 & 20 & 32 & 36 & 28 & 24 & 24 & 28 & 32 & 20 & 40 & 24 & 4 & 44 & 12 \\
claude-haiku-4.5 & 40 & 28 & 48 & 12 & 16 & 40 & 32 & 32 & 48 & 20 & 32 & 20 & 20 & 36 & 24 & 24 & 12 & 0 & 20 & 8 \\
gemini-2.5-flash & 16 & 12 & 0 & 0 & 0 & 12 & 0 & 12 & 4 & 4 & 12 & 12 & 4 & 24 & 4 & 8 & 16 & 0 & 16 & 8 \\
claude-haiku-3.5 & 8 & 8 & 8 & 4 & 8 & 0 & 4 & 16 & 8 & 16 & 8 & 12 & 8 & 8 & 0 & 4 & 8 & 0 & 8 & 4 \\
\addlinespace[0.1cm]
\midrule
\multicolumn{21}{l}{\textit{Open-source models}} \\
\midrule
deepseek-chat-v3.1 & 40 & 24 & 44 & 28 & 36 & 12 & 4 & 20 & 24 & 20 & 16 & 4 & 20 & 24 & 32 & 16 & 20 & 8 & 8 & 40 \\
qwen3-next-80b-a3b & 16 & 28 & 8 & 24 & 12 & 24 & 24 & 12 & 36 & 24 & 12 & 12 & 12 & 16 & 12 & 8 & 20 & 8 & 24 & 20 \\
deepseek-r1-0528 & 16 & 20 & 4 & 24 & 24 & 24 & 12 & 16 & 32 & 36 & 12 & 4 & 12 & 20 & 8 & 8 & 16 & 8 & 4 & 20 \\
qwen3-max & 16 & 20 & 4 & 12 & 8 & 16 & 8 & 12 & 24 & 12 & 16 & 40 & 8 & 20 & 12 & 8 & 20 & 0 & 8 & 12 \\
llama-4-maverick & 24 & 0 & 4 & 8 & 8 & 4 & 4 & 16 & 12 & 12 & 12 & 8 & 16 & 4 & 4 & 0 & 12 & 0 & 12 & 12 \\
mistral-medium-3.1 & 12 & 8 & 0 & 4 & 8 & 8 & 12 & 8 & 16 & 0 & 8 & 12 & 8 & 8 & 12 & 4 & 12 & 0 & 12 & 8 \\
llama-4-scout & 0 & 0 & 0 & 0 & 4 & 0 & 0 & 0 & 0 & 0 & 0 & 0 & 0 & 0 & 0 & 0 & 0 & 0 & 0 & 4 \\
gemma-3-27b & 0 & 0 & 0 & 0 & 0 & 0 & 0 & 0 & 0 & 0 & 0 & 0 & 0 & 0 & 0 & 0 & 0 & 0 & 0 & 0 \\
\midrule
\addlinespace[0.1cm]
\multicolumn{21}{l}{\textit{Ours}} \\
\midrule
C1-SFT-4B & 48 & 52 & 72 & 28 & 48 & 36 & 28 & 52 & 48 & 52 & 56 & 40 & 24 & 48 & 60 & 44 & 44 & 12 & 60 & 72 \\
C1-SFT-8B & 52 & 44 & 84 & 20 & 60 & 40 & 56 & 44 & 40 & 52 & 64 & 40 & 36 & 48 & 60 & 36 & 44 & 8 & 48 & 56 \\
\rowcolor{gray!20} C1-4B & 64 & 64 & 84 & 56 & 60 & 36 & 44 & 52 & 36 & 64 & 64 & 56 & 28 & 52 & 68 & 60 & 52 & 4 & 52 & 76 \\
\bottomrule
\end{tabular}}
\end{table*}

\begin{algorithm}[t]
\caption{Theme-Balanced Data Sampling}
\label{alg:balanced}
\begin{algorithmic}
\small
\REQUIRE Dataset $\mathcal{D}$ where each puzzle $p$ has theme set $\mathcal{T}(p)$
\REQUIRE Number of rare themes to balance $K$
\REQUIRE Maximum samples per theme $M$
\ENSURE Balanced subset $\mathcal{D}_{\text{bal}}$
\STATE Compute theme frequencies: $f(t) \gets |\{p \in \mathcal{D} : t \in \mathcal{T}(p)\}|$ for all themes $t$
\STATE Select rare themes: $\mathcal{T}_{\text{rare}} \gets \arg\min_K f(t)$
\STATE Initialize selected IDs: $\mathcal{S} \gets \emptyset$
\STATE Initialize output: $\mathcal{D}_{\text{bal}} \gets \emptyset$
\FOR{each theme $t \in \mathcal{T}_{\text{rare}}$}
    \STATE $\mathcal{C}_t \gets \{p \in \mathcal{D} : t \in \mathcal{T}(p) \land \text{id}(p) \notin \mathcal{S}\}$
    \STATE Sample $\min(M, |\mathcal{C}_t|)$ puzzles from $\mathcal{C}_t$ without replacement
    \STATE $\mathcal{D}_{\text{bal}} \gets \mathcal{D}_{\text{bal}} \cup \text{sampled puzzles}$
    \STATE $\mathcal{S} \gets \mathcal{S} \cup \{\text{id}(p) : p \in \text{sampled puzzles}\}$
\ENDFOR
\STATE \textbf{return} $\mathcal{D}_{\text{bal}}$
\end{algorithmic}
\end{algorithm}


\end{document}